\documentclass{article}
\usepackage[usenames,dvipsnames]{color}
\usepackage[preprint]{spconf}
\usepackage{amsmath,graphicx,lipsum}
\usepackage{todonotes}
\usepackage{multirow}

\usepackage{mathtools}
\DeclarePairedDelimiter\floor{\lfloor}{\rfloor}

\title{mask-combine decoding and classification approach for punctuation prediction with real-time inference constraints}
\name{Christoph Minixhofer, Ondřej Klejch, Peter Bell}
\address{Centre for Speech Technology Research, University of Edinburgh, Edinburgh EH8 9AB, UK\\
\texttt{\{christoph.minixhofer,o.klejch,peter.bell\}@ed.ac.uk}}
\copyrightnotice{\copyright~IEEE 2021.}

\begin{document}
%
\maketitle
\begin{abstract}
In this work, we unify several existing decoding strategies for punctuation prediction in one framework and introduce a novel strategy which utilises multiple predictions at each word across different windows. We show that significant improvements can be achieved by optimising these strategies after training a model, only leading to a potential increase in inference time, with no requirement for retraining. We further use our decoding strategy framework for the first comparison of tagging and classification approaches for punctuation prediction in a real-time setting. Our results show that a classification approach for punctuation prediction can be beneficial when little or no right-side context is available.
\end{abstract}
\section{Introduction}
\label{sec:intro}

We consider the task of automatically punctuating the output of an automatic speech recognition (ASR) system.  Most modern ASR systems produce output without punctuation, while many downstream tasks commonly used with ASR -- such as machine translation, sentiment classification or intent recognition -- expect punctuated input. For some applications, punctuation prediction has to be performed in real-time. Notable examples include dictation systems and captioning of TV broadcasts.  When the aforementioned downstream tasks have a real-time requirement on their own, real-time punctuation prediction is required as well. The currently prevalent approach for punctuation prediction is a sequence tagging one, where for each word in a window, a punctuation class (tag) is predicted \cite{yi2020adversarial,chen2020controllable,Lin2020disf,sotapunctuation}. Although these models are extensively evaluated, little information exists on their performance when there are real-time constraints in the form of limited right-side context. Additionally, different models use different decoding strategies for conducting inference, making comparison difficult.
In this work, we introduce mask-combine decoding, which a) can be used to impose real-time constraints on current state-of-the-art models used for punctuation prediction utilising tagging approaches b) unifies several decoding approaches in one framework, allowing decoding to be treated as a set of hyper-parameters c) introduces the combination of overlapping probability distributions which leads to an incremental improvement when compared to previous decoding strategies. We obtain results using the aforementioned tagging approach prevalent in previous work \cite{yi2020adversarial,chen2020controllable,Lin2020disf,sotapunctuation}, but use our decoding strategy with parameters limiting right-side context to simulate a real-time use case. This makes it possible to compare previous techniques with a novel sequence classification approach for punctuation prediction, in which only one punctuation mark at a specific location with limited lookahead is predicted for each sequence. For this, we introduce a special \texttt{[PUNCT]} token and use this approach to emphasise samples with limited right-side context during training.

\section{Related Work}
\label{sec:format}
While earlier approaches such as hidden-markov-models or finite-state machines model punctuation as a probability distribution over possible events occurring between words, recent punctuation prediction methods predominantly use neural-network-based models. The tasks used for training these models are as follows:
\begin{itemize}
    \item \textit{machine translation} task where, using an input sequence without punctuation, an output sequence of punctuation marks (or text including punctuation marks) is predicted \cite{Klejch2016,Klejch2017,yi2019speech2vec}
    \item \textit{sequence tagging} task in which for each input, a probability distribution across possible punctuation marks is predicted \cite{yi2020adversarial,chen2020controllable,Lin2020disf,sotapunctuation,Tilk2015}
    \item \textit{sequence classification} task in which for each sequence, a probability distribution across possible punctuation marks for \textbf{a fixed location} within the sequence is predicted \cite{che2016}
\end{itemize}
Recent work on punctuation prediction relies on transformer \cite{vaswani2017} models pre-trained on large corpora in an unsupervised way. These approaches frequently train and evaluate their models on the IWSLT11 dataset \cite{iwslt2011} which is processed as an unsegmented transcript \cite{che2016}. Recent models achieving state-of-the-art results add a classification head to a pre-trained transformer \cite{sotapunctuation,chen21d_interspeech} and fine-tune on the IWSLT11 train dataset using a sequence tagging task. We hypothesise the sequence tagging, rather than classification, is used because the tagging approach can predict $w$ punctuation marks at once, where $w$ is the window size used. For the classification approach, inference has to be conducted once for each word. While this is a downside when long transcripts have to be processed quickly, this is not the case for a real-time setting. The closer the target latency is to 0 words, the more often inference has to be conducted, making the tagging approach similar to the classification one in efficiency. There is recent work on punctuation prediction with a focus on the real-time use case, but with a tagging rather than classification approach \cite{nguyen2019,chen2020controllable}. Nguyen et al. \cite{nguyen2019} create a model for fast punctuation prediction, with a latency of 20 words \cite{chen2020controllable}. Chen et al. introduce a controllable time-delay transformer \cite{chen2020controllable} with a latency of 10 words and comparable performance. Chen et al. \cite{chen21d_interspeech} note that using this approach, performance at the beginning and end of these windows is degraded due to the lack of right or left-side context, and introduce a "Double-Overlap Sliding Window Decoding Strategy", in which a fixed number of words on the left and right side of each window are masked out and not used for prediction. This is inspired by Nguyen et al. \cite{nguyen2019}, who use overlapping chunks and set a boundary within each set of overlapping chunks to determine which predictions are used — they report a relative improvement of $\approx1\%$ F1 for different punctuation classes using this decoding strategy. Cho et al. \cite{Cho2012} use a similar approach by using overlapping windows and only inserting punctuation if a punctuation mark appears more frequently than a set threshold. As they use a machine translation approach, their models performance is measured in BLEU, on which they achieve an improvement of $\approx1\%$ as well. No previous work has utilised the probability distribution across punctuation marks across different windows. Existing approaches either drop predictions without right or left-side context in favour of ones with more context \cite{nguyen2019,chen21d_interspeech}, or use the number of predicted marks when several predictions are made for one token as a threshold \cite{Cho2012}. To the best of our knowledge, there are no sequence classification approaches that target real-time punctuation prediction.

\section{Mask-Combine Decoding}
\label{sec:methods}

\begin{figure}[tb]
\begin{minipage}[b]{1.0\linewidth}
  \centering
  \centerline{\includegraphics[width=8cm]{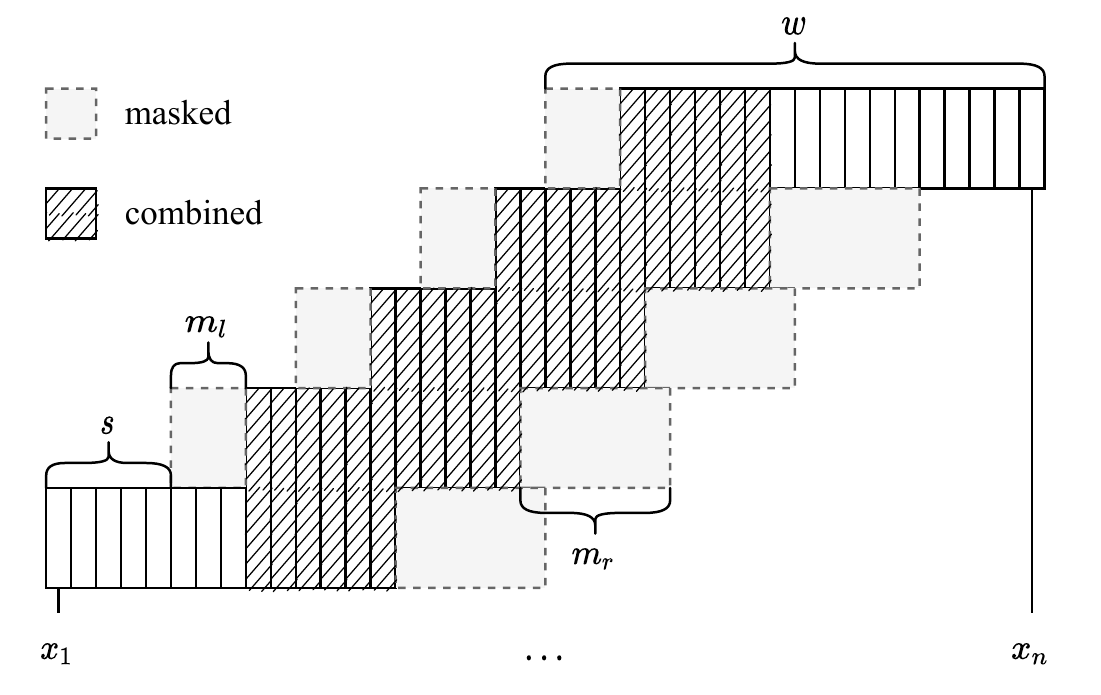}}\medskip
\end{minipage}
\caption{Mask-combine decoding with stride $s=s_2=5$, left/right mask $m_l=3$/$m_r=6$ and window size $w=20$.}
\label{fig:maskedmean}
\end{figure}
We introduce the \textit{mask-combine} decoding strategy, which uses overlapping sliding windows over the unsegmented transcript of words, as can be seen in Figure~\ref{fig:maskedmean}, extending the double-overlap sliding window decoding strategy \cite{chen21d_interspeech}. A number of left and right predictions (determined by $m_l$ and $m_r$ respectively) of each window are masked and not taken into consideration. We then allow the stride $s$ to be set separately from $m_l$ and $m_r$, resulting in multiple probability distributions from different windows at each word. The stride length with overlap $n$ of $1$ (no overlap) is defined as $w-(m_l+m_r)$, which we call $s_1$.
To obtain $n$ predictions at each word, we then divide $s_1$ by $n$, obtaining $s_n$.
$$s_n=\floor*{\frac{1}{n}({w-(m_l+m_r)})}$$
If the original stride length is not divisible by $n$, we can use the floor of the result to ensure we have at least $n$ probability distributions at each step, with more than $n$ at some steps. This way we can obtain $n$ probability distributions for any $s_n$ with $n\leq s_1$ for every word after the first $s+m_l$ and before the last $s+m_r$ words in the unsegmented transcript. While previous work combined predictions using a vote \cite{Cho2012}. We use the resulting overlapping probability distributions. These can be combined using various methods. In this work, we use the mean of all probability distributions at a time step as a baseline. We then investigate an entropy-score-weighed sum \cite{tornetta2021entropy} to account for probability distributions with low confidence, and a hamming window to account for predictions with less left or right-side context performing worse.
The parametric nature of the mask-combine strategy allows both double-overlap sliding window decoding \cite{chen21d_interspeech} and overlapped-chunk split and merging algorithm \cite{nguyen2019} to be used. For double-overlap sliding window decoding we set $s=w-(m_l+m_r)$ and leave the left and right mask unchanged. For the overlapped-chunk split and merging algorithm, we set $m_r=\text{min\_words\_cut}$, $m_l=\text{overlap\_size}-\text{min\_words\_cut}$ (see Nguyen et al. \cite{nguyen2019} for \texttt{min\_words\_cut} and \texttt{overlap\_size}), while leaving $s$ unchanged.
\section{Classification Approach}
\begin{figure}[tb]
\begin{minipage}[b]{1.0\linewidth}
  \centering
  \centerline{\includegraphics[width=8cm]{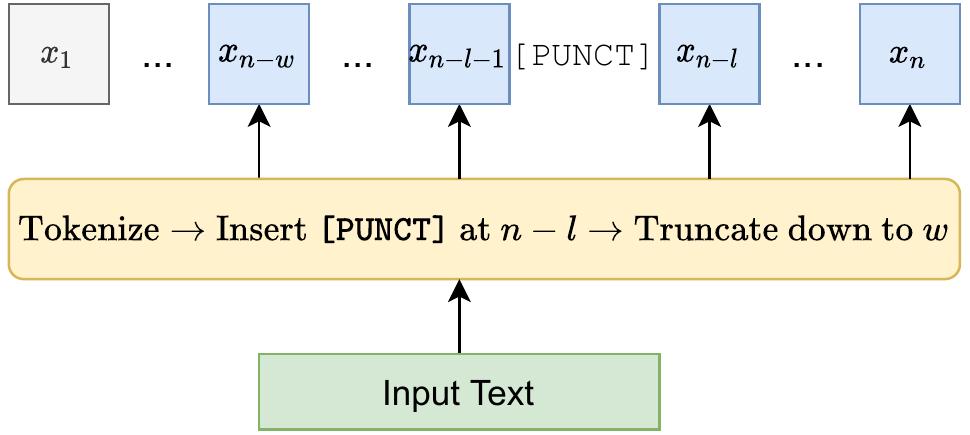}}\medskip
\end{minipage}
\caption{Tokenization, \texttt{[PUNCT]} token insertion and truncation.}
\label{fig:tokinstrunc}
\end{figure}

For real-time punctuation prediction, little right-side context is available. It could therefore be beneficial to train a model with emphasis on word with little such context. The current tagging approaches most widely used only have a few words with little right-side context per training sample. To counteract this, we train a model with explicitly reduced right-side context. We do this by changing the task to a sequence classification one. To retain some flexibility over the amount of right-side context, we additionally introduce a \texttt{[PUNCT]} token which informs the model of the position of the punctuation to be predicted. As can be seen in Figure~\ref{fig:tokinstrunc}, the input text is tokenized, the punctuation token is inserted at $n$ (the number of tokens) subtracted by the lookahead $l$. As a last step, the remaining tokens are truncated (removing tokens from the left) until $w$ tokens remain.
Another feature of utilising a decoding strategy with $m_l$ and $m_r$ parameters such as the mask-combine decoding introduced in this work or \textit{double-overlap sliding window decoding} is the possibility to compare the currently predominant sequence tagging approach, in which a punctuation token is predicted for each word in the input sequence, with the classification approach outlined above. To compare the two at a right-side context of $l$, we can set $m_l=s-l-1$ and $m_r=l$.

The dot-product attention used in transformers scales quadratically with sequence length ($\mathcal{O}(n^2)$) \cite{tay2020efficient}, which means reducing the window size $w$ can lead to a significant impact, which is beneficial for both training and fast inference. With this classification approach, the window size $w$, in combination with the stride $s$ can be used as hyper-parameters to control the total amount of data (a higher $s$ will lower the number of samples) and the time needed to process each sample (a higher $w$ will increase the training complexity at each sample).

\section{Experiments}

\begin{figure}[tb]
\begin{minipage}[b]{1.0\linewidth}
  \centering
  \centerline{\includegraphics[width=7.5cm]{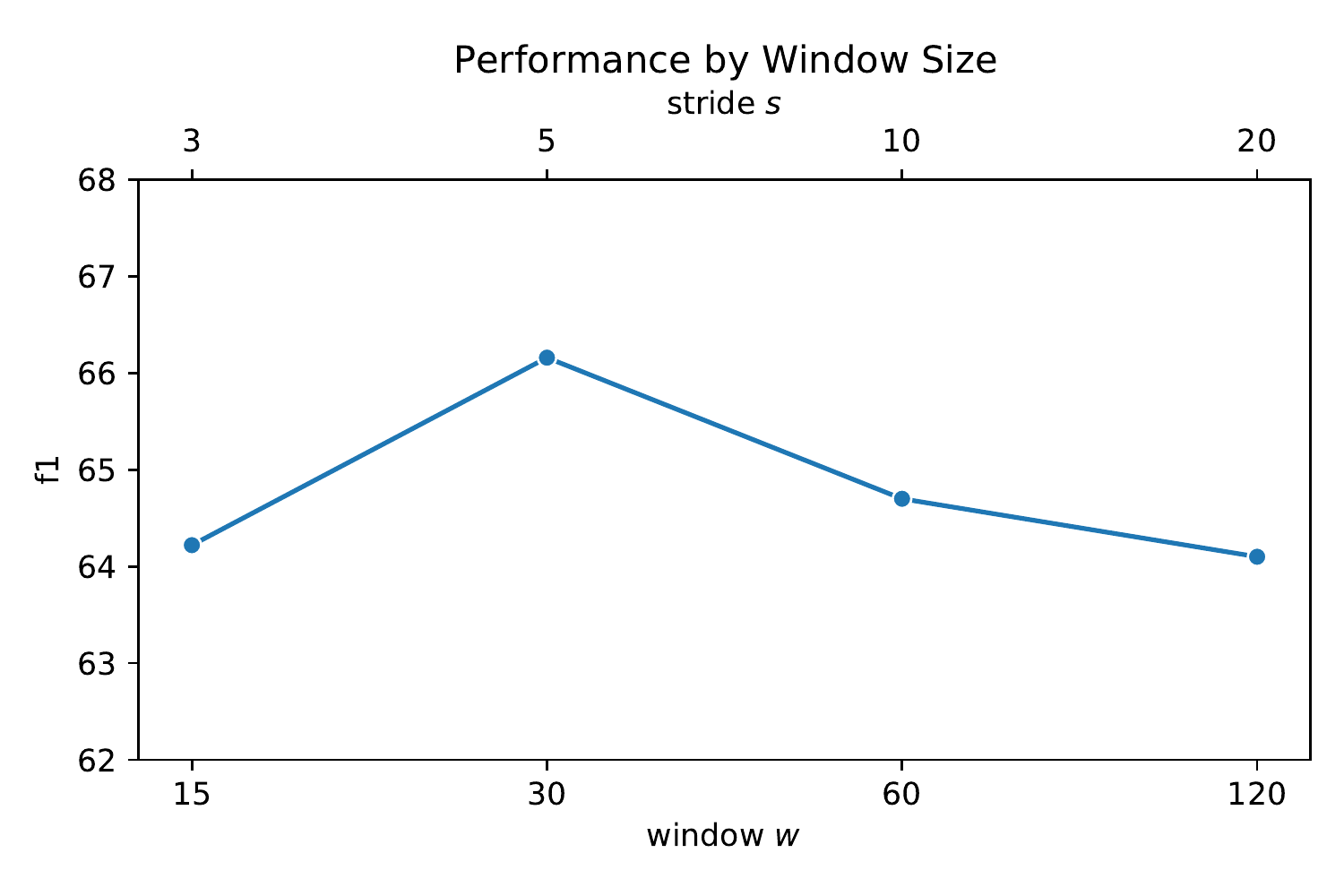}}\medskip
\end{minipage}
\caption{Performance with different window and stride sizes, given the same compute budget.}
\label{fig:windowstride}
\end{figure}

\begin{table*}[]
\centering
\begin{tabular}{l|c|rcc|ccc|ccc|ccc}
\hline
\multicolumn{1}{c|}{\multirow{2}{*}{\textbf{Task}}} &
  \multirow{2}{*}{\textbf{$l$}} &
  \multicolumn{3}{c|}{\textbf{Comma}} &
  \multicolumn{3}{c}{\textbf{Period}} &
  \multicolumn{3}{c|}{\textbf{Question}} &
  \multicolumn{3}{c}{\textbf{Overall}} \\ \cline{3-14} 
\multicolumn{1}{c|}{}           &   & \multicolumn{1}{c}{P}    & R    & $F_1$ & P    & R    & $F_1$ & P    & R    & $F_1$ & P    & R    & $F_1$         \\ \hline
\multirow{5}{*}{Tagging}        & 0 & 49.2                     & 30.7 & 37.8  & 46.7 & 43.7 & 45.2  & 57.9 & 47.8 & 52.4  & 48.1 & 37.4 & 42.1          \\
                                & 1 & 63.1                     & 62.3 & 62.7  & 68.4 & 72.0 & 70.2  & 66.1 & 80.4 & 72.5  & 65.8 & 67.5 & \textbf{66.6} \\
                                & 2 & 68.0                     & 68.6 & 68.3  & 75.1 & 78.1 & 76.6  & 73.1 & 82.6 & 77.6  & 71.6 & 73.5 & \textbf{72.5} \\
                                & 3 & 70.4                     & 71.8 & 71.1  & 78.9 & 82.0 & 80.4  & 70.4 & 82.6 & 76.0  & 74.5 & 77.0 & \textbf{75.7} \\
                                & 4 & 69.9                     & 71.7 & 70.8  & 80.4 & 83.4 & 81.8  & 74.5 & 82.6 & 78.4  & 75.1 & 77.6 & \textbf{76.3} \\ \hline
\multirow{5}{*}{Classification} & 0 & 49.3                     & 34.7 & 40.8  & 47.7 & 57.7 & 52.2  & 42.9 & 65.2 & 51.7  & 48.1 & 46.6 & \textbf{47.3} \\
                                & 1 & 61.2                     & 60.7 & 60.9  & 65.6 & 74.2 & 69.6  & 57.1 & 78.3 & 66.1  & 63.3 & 67.6 & 65.4          \\
                                & 2 & 64.5                     & 66.6 & 65.5  & 74.9 & 78.7 & 76.7  & 65.5 & 78.3 & 71.3  & 69.5 & 72.7 & 71.1          \\
                                & 3 & 67.2                     & 67.8 & 67.5  & 76.7 & 81.1 & 78.9  & 63.8 & 80.4 & 71.2  & 71.7 & 74.5 & 73.1          \\
                                & 4 & \multicolumn{1}{c}{66.8} & 68.6 & 67.7  & 78.3 & 82.1 & 80.2  & 66.1 & 84.8 & 74.3  & 72.3 & 75.6 & 73.9          \\ \hline
\end{tabular}
\caption{$F_1$ (\%) values for classification and tagging approaches for real-time punctuation prediction with varying lookahead $l$.}
\label{tab:classtag}
\end{table*}

For this work, we conduct experiments on a BERT-base \cite{devlin2018} model finetuned on the IWSLT11 train dataset using the hyperparameters found by previous work \cite{chen21d_interspeech}. To allow for comparison with previous work, we use reference transcriptions rather than ASR output.
As Chen et al. \cite{chen21d_interspeech} do not report the improvement of their overlapping decoding strategy over a non-overlapping one, we first replicate their results without their strategy, which is equivalent to the \textit{mask-combine decoding} we introduced with parameters $s=120, m_l=0, m_r=0$ (unmasked), which results in $n=1$ (one prediction per word). We repeat this step with their strategy ($s=70, m_l=30, m_r=15$, $n=1$) (masked). As can be seen in see Table~\ref{tab:mask}, we then conduct several experiments using masked-mean decoding resulting in $n=2$ and $n=4$ by reducing the step size as described in Section~\ref{sec:methods} and combine the probability distributions at each step using their mean.
We also test weighing predictions at each step using a hamming window and entropy score (as a proxy for confidence) respectively, but neither improve performance. Overall, using overlapping predictions leads to a minor improvement of \textbf{0.4\%} $F_1$ (absolute) when using the same mask as in \cite{chen21d_interspeech}. When not using a mask, the improvement is \textbf{1.7\%} $F_1$ (absolute). This is to be expected, as the words with little right or left-side context, which are otherwise masked out, benefit the most from the ensembling effect of overlapping prediction windows.

We next evaluate which window size is optimal for a classification approach given the same compute budget as the sequence tagging approach. We use the previously described \texttt{[PUNCT]} token and lookahead (right-context length) values ranging from 0 to 4. We only train the model for one epoch and balance the window size $w$ and stride $s$ parameters, with the former reducing inference time while potentially lowering prediction performance (when too little left-side context is available), and the latter increasing inference time while increasing prediction performance due to a higher number of seen samples. We assess performance by computing the average of $F_1$ at the different lookahead values. As can be seen in Figure~\ref{fig:windowstride}, a window size as low as 15 decreases performance, while after increasing beyond a window size of 60, performance decreases due to less training data. Therefore, out of the values tested, we find $w=30,s=5$ to perform the best and use this configuration for the subsequent experiments.

Our intuition being that the previously trained classification model could be better suited to real-time punctuation prediction than the tagging model, we evaluate both at different lookahead values. On the tagging model, we achieve this by setting the parameters of our previously introduced mask-combine decoding to $s=1,m_l=w-l-1,m_r=l$. As can be seen in Table~\ref{tab:classtag}, the classification approach brings a \textbf{5.2\%} $F_1$ (absolute) improvement when using it for real-time punctuation prediction with no given lookahead. At lookaheads greater than one, the classification approach is on average \textbf{1.9\%} $F_1$ (absolute) lower than the tagging method. We reason that this is due to the tagging method encountering more punctuation marks during training due to the nature of the task, outweighing the potential benefit of the classification method encountering proportionally more punctuation marks with little right-side context.

\begin{table}[]
\begin{tabular}{lccc}
\multicolumn{1}{c}{\multirow{2}{*}{\textbf{Model}}} & \multicolumn{3}{c}{\textbf{Overall}}                        \\
\multicolumn{1}{c}{}                                & P                        & R                        & $F_1$ \\ \hline
BERT-base (baseline) [9]                            & 76.3                     & 78.4                     & 77.4  \\ \hline
BERT-base (ours)                                    &                          &                          &       \\
\textcolor{red}{without} mask, $n=1$                      & 75.1                     & 78.0                     & 76.5  \\
\textcolor{red}{without} mask, $n=2$                      & 75.9                     & 78.5                     & 77.2  \\
\textcolor{red}{without} mask, $n=4$                      & 76.9                     & 79.5                     & 78.2  \\
\textcolor{PineGreen}{with} mask, $n=1$                    & 76.4                     & 79.6                     & 78.0  \\
\textcolor{PineGreen}{with} mask, $n=2$                    & \multicolumn{1}{l}{76.6} & \multicolumn{1}{l}{80.0} & 78.2  \\
\textcolor{PineGreen}{with} mask, $n=4$                    & \multicolumn{1}{l}{76.8} & \multicolumn{1}{l}{80.1} & \textbf{78.4} 
\end{tabular}
\caption{Comparison of decoding strategy with and without mask and different overlap $n$ and previous work.}
\label{tab:mask}
\end{table}

\section{Conclusion}
We introduce the mask-combine decoding strategy for punctuation prediction, which builds upon previous methods and extends them by adding the possibility of combining overlapping predictions. The previous double-overlap sliding window decoding \cite{chen21d_interspeech} and overlapped-chunk split and merging algorithm \cite{nguyen2019} can be expressed by setting the parameters of mask-combine decoding to specific values. Future work could investigate extending mask-combine decoding to be able to express the fast decoding strategy used used with the CT-Transformer \cite{chen2020controllable}. We find that choosing a well-optimised decoding strategy alone can lead to significant improvements, while not necessitating computationally expensive retraining of models. We find that the decoding strategy introduced by Chen et al. \cite{chen21d_interspeech}, combined with addition overlapping probability distributions with $n=4$ yields and absolute improvement over a baseline using no particular decoding strategy of 1.9\% (absolute) $F_1$, which is comparable to the improvement reported when using self-training with a large (30M words) unlabeled corpus, which is reported to yield a 2.2\% improvement \cite{chen21d_interspeech}.
We further use our decoding strategy to conduct a comparison between the commonly used sequence tagging approach for punctuation prediction, and a novel classification approach utilising a \texttt{[PUNCT]} token at varying positions with little right-side context. To allow for as many samples to be used for training as possible, while conducting a fair comparison by keeping a similar computing budget, we optimise the window size and stride length for the classification approach. Ultimately, we show there is a latency-accuracy trade-off between the two models, with the classification model performing significantly better when there is no future context. Conversely, the tagging model outperforms the classification model slightly as soon as there is any right-side context available.

As our classification approach aims to improve performance in a real-time setting with little lookahead, such as streamed (automatic speech recognition) ASR output, future work could investigate the robustness of our approach to ASR errors. Decoding the punctuation in a real-time setting could be investigated as well, for example by dynamically varying lookahead based on prediction confidence.
\section{Acknowledgements}
Research supported with Cloud TPUs from Google’s
Tensorflow Research Cloud (TRC).


\vfill\pagebreak

\bibliographystyle{IEEEbib}
\bibliography{refs}

\end{document}